\documentclass[runningheads]{llncs}
\pdfoutput=1 
\usepackage[dvipsnames]{xcolor} 
\usepackage{graphicx}
\usepackage{comment}
\usepackage{amsmath,amssymb} 
\usepackage{color}
\usepackage{ulem}

\usepackage{epsfig}
\usepackage{graphicx}
\usepackage{amsmath}
\usepackage{amssymb}
\usepackage{amsfonts}

\usepackage{cellspace}
\usepackage{bm}
\usepackage{physics} 
\usepackage[protrusion=true,tracking=true,kerning=true,expansion,final]{microtype}      
\usepackage[dvipsnames]{xcolor} 
\usepackage{booktabs} 
\usepackage{subcaption}
\usepackage{nicefrac} 
\usepackage{cancel}
\usepackage[round, compress]{natbib}

\renewcommand{\cite}[1]{\citep{#1}}
\usepackage{xspace}
\usepackage{upref}

\usepackage{gensymb}
\usepackage{bbding}
\usepackage{mathtools}
\usepackage{centernot} 
\usepackage{bigints}
\usepackage{dsfont} 
\usepackage{esint} 
\usepackage{authblk}
\usepackage{multirow}
\usepackage{enumitem}

\usepackage{algorithm}
\usepackage{algpseudocode}
\usepackage{algorithmicx}

\usepackage[scaled=.98,sups,osf]{XCharter}
\usepackage[libertine,vvarbb,scaled=1.07]{newtxmath}

\usepackage{varioref} 
\usepackage[pagebackref=true,breaklinks=true,letterpaper=true,colorlinks,bookmarks=True]{hyperref}
\usepackage[capitalise]{cleveref} 
\crefname{appsec}{appendix}{appendices}
\Crefname{appsec}{Appendix}{Appendices}

\usepackage{url}

\definecolor{mydarkblue}{rgb}{0,0.08,0.45}
\definecolor{urlcolor}{rgb}{0,.145,.698}
\definecolor{linkcolor}{rgb}{.71,0.21,0.01}
\hypersetup{ %
	pdftitle={Self-Supervised Learning for Large-Scale Unsupervised Image Clustering},
	pdfauthor={Evgenii Zheltonozhskii, Chaim Baskin, Alex M. Bronstein, Avi Mendelson},
	pdfsubject={},  
	pdfkeywords={},
	pdfborder=0 0 0,     
    pdfpagemode=UseOutlines,
	breaklinks=true,  
	colorlinks=true,
	letterpaper=true,
	bookmarks=true,
	bookmarksnumbered=true,
	urlcolor=urlcolor,
	linkcolor=linkcolor,
	citecolor=mydarkblue,
	filecolor=mydarkblue,
	pdfview=FitH}

\renewcommand*{\backref}[1]{} 
\renewcommand*{\backrefalt}[4]{%
	\ifcase #1 %
	\or
	(cited on p. #2)%
	\else
	(cited on pp. #2)%
	\fi
}
\makeatletter
\renewcommand{\@biblabel}[1]{#1.}
\makeatother

\vbadness=\maxdimen
\usepackage{trajan}
\usepackage{docmute} 
\usepackage{xr}

\renewcommand{\emph}{\textit}

\usepackage{bm}

\begin{document}
\pagestyle{headings}
\mainmatter
\def\ECCVSubNumber{7380}  

\title{Self-Supervised Learning for Large-Scale Unsupervised Image Clustering} 

\titlerunning{Self-Supervised Learning for Large-Scale Unsupervised Image Clustering}
%

\newcommand*\samethanks[1][\value{footnote}]{\footnotemark[#1]}

\author{Evgenii Zheltonozhskii\inst{1}\orcidID{0000-0002-5400-9321} \and Chaim Baskin\inst{1} \and Alex M. Bronstein\inst{1} \and Avi Mendelson\inst{1}}
\authorrunning{E. Zheltonozhskii et al.}
%
\institute{
Technion -- Israel Institute of Technology \\
\email{
\href{mailto:evgeniizh@campus.technion.ac.il}{evgeniizh@campus.technion.ac.il};
\href{mailto:chaimbaskin@campus.technion.ac.il}{chaimbaskin@campus.technion.ac.il};
\href{mailto:bron@cs.technion.ac.il}{bron@cs.technion.ac.il};
\href{mailto:avi.mendelson@cs.technion.ac.il}{avi.mendelson@cs.technion.ac.il} } }

{\maketitle}

\begin{abstract}
Unsupervised learning has always been appealing to machine learning researchers and practitioners, allowing them to avoid an expensive and complicated process of labeling the data. However, unsupervised learning of complex data is challenging, and even the best approaches show much weaker performance than their supervised counterparts.  
Self-supervised deep learning has become a strong instrument for representation learning in computer vision. However, those methods have not been evaluated in a fully unsupervised setting. In this paper, we propose a simple scheme for unsupervised classification based on self-supervised representations. We evaluate the proposed approach with several recent self-supervised methods showing that it achieves competitive results for ImageNet classification ($39\%$ accuracy on ImageNet with $1000$ clusters and $46\%$ with overclustering). We suggest adding the unsupervised evaluation to a set of standard benchmarks for self-supervised learning.  The code is available at \url{https://github.com/Randl/kmeans_selfsuper}.
\end{abstract}


\section{Introduction}
\label{sec:intro}
Deep learning has become the primary tool in various computer vision tasks, being especially successful in image classification, detection, and segmentation. However, along with massive computing resources required to train state-of-the-art neural networks (NNs), massive datasets with millions of labeled samples are a necessary part of its success. Since creating those datasets is a costly procedure, researchers have recently started looking at the methods of training NNs without labeled data. Those methods commonly referred to as \emph{self-supervised learning} recently has become a powerful instrument for large-scale computer vision.

A result of training a network in a self-supervised manner is usually a representation: a vector in a latent space. Evaluation of those representations proceeds mainly by two following approaches: fine-tuning the network as a feature extractor for some task
(common choices are segmentation tasks or ImageNert classification on a small amount of data, e.g., $1\%$ of labels)
or training a linear classifier on the extracted features. A variation of the latter is to train a $k$-nearest neighbor classifier instead of a linear classifier. While linear classification directly evaluates the learned representation, it is not always capable of predicting performance on downstream tasks \cite{resnick2019probing}. On the other hand, performance of a fine-tuned network strongly depends on the training procedure which is hard to separate from the quality of the representation itself.  


In computer vision, the main approaches to training a network in a self-supervised manner are contrastive losses, pretext tasks, and generative models.
Contrastive methods \cite{oord2018cpc,ye2019invariant,ermolov2020whitening} try to create different views of the same image and bring a representation of different views closer and representations of different images farther apart.
Alternatively, it is possible to train the network to perform some label-free pretext task, such as predicting  context \cite{doersch2015context}, image  rotation \cite{gidaris2018unsupervised,kolesnikov2019revisiting}, colorization \cite{zhang2016colorful}, solving ``jigsaw puzzle'' \cite{kim2018learning}, etc.
Generative-based representation learning uses the latent vectors of a generative model, e.g., Boltzmann machines \cite{lee2009convolutional}, autoencoders \cite{caron2018clustering} or GANs \cite{donahue2016bigan, donahue2019bigbigan}, as a representation.

The overview of recent methods of self-supervised and unsupervised approaches is given in \cref{tab:ss-links}.
\begin{table}
\centering
\caption{Brief review of selected self-supervised methods, including links to code and linear-evaluation performance. Clustering from pretext \cite{gansbeke2020scan} is an unsupervised method. First plus in each row is a  hyperlink. }
\label{tab:ss-links}
\begin{tabular}{lcccc}
\toprule
\textbf{Method}   & \textbf{Code} & \textbf{Checkpoints} & \textbf{ResNet-50}&\textbf{Best}\\ \midrule
CPC \cite{oord2018cpc}&--&--&--&$48.7\%$\\ 
CPC v2 \cite{henaff2019cpc2}&--&--&$63.8\%$&$71.5\%$\\ 
AMDIM \cite{bachman2019amdim}&\href{https://github.com/Philip-Bachman/amdim-public}{+}&+&--&$68.1\%$\\ 
CMC \cite{tian2019contrastive}&\href{https://github.com/HobbitLong/CMC/}{+}&+&$66.2\%$&$70.6\%$\\ 
BigBiGAN \cite{donahue2019bigbigan}&--&\href{https://tfhub.dev/s?publisher=deepmind&q=bigbigan}{+}&--&$61.3\%$\\ 
MoCo \cite{he2019moco}&\href{https://github.com/facebookresearch/moco}{+}&+&$60.6\%$&$68.6\%$\\ 
Self-Label \cite{asano2019selflabelling}&\href{https://github.com/yukimasano/self-label}{+}&+&$71.1\%$&$71.1\%$\\ 
SimCLR \cite{chen2020simclr}&\href{https://github.com/google-research/simclr}{+}&+&$69.3\%$&$76.5\%$\\ 
MoCo v2 \cite{chen2020moco2}&\href{https://github.com/facebookresearch/moco}{+}&+&$71.1\%$&$71.1\%$\\ 
InfoMin \cite{tian2020infomin}&\href{https://github.com/HobbitLong/PyContrast}{+}&+&$73.0\%$&$75.2\%$\\ 
BYOL \cite{grill2020byol}&\href{https://github.com/deepmind/deepmind-research/tree/master/byol}{+}&+&$74.3\%$&$79.6\%$\\ 
SwAV \cite{caron2020swav}&\href{https://github.com/facebookresearch/swav}{+}&+&$75.3\%$&$78.5\%$\\ 
SimCLRv2 \cite{chen2020simclr2}&\href{https://github.com/google-research/simclr}{+}&+&$71.7\%$&$79.8\%$\\ 
iGPT \cite{chen2020igpt}&\href{https://github.com/openai/image-gpt}{+}&+&$-$&$72.0\%$\\ \midrule
Clustering from pretext \cite{gansbeke2020scan}&\href{https://github.com/wvangansbeke/Unsupervised-Classification}{+}&+&--&--\\
 \bottomrule
%
\end{tabular}
\end{table}

\paragraph{Contribution}
In this paper, we propose an additional way of evaluating self-supervised learning: training a clustering algorithm on extracted features in an unsupervised manner. While this method suffers from similar disadvantages as linear evaluation, it can provide additional insights and a benchmark for unsupervised learning on large-scale datasets, such as ImageNet.  We also show that self-supervised learning provides a strong baseline for unsupervised computer vision and mentions some possible direction for the current self-supervised methods performance improvement.

Thanks to the increasing trend of publishing pre-trained models and code, we were able to test the existing approaches on the proposed benchmark. In particular, we show that the best-performing self-supervised algorithm achieves almost $40\%$ top-1 accuracy on ImageNet without any supervision. Those results are on par with a specialized clustering approach by \citet{gansbeke2020scan}. We also evaluate ObjectNet \cite{barbu2019objectnet}, a dataset created for testing image classification algorithms in conditions closer to real-life, and conclude that it is hard to achieve generalization in unsupervised settings. 

This benchmark provides a more challenging task for future self-supervised learning approaches, allowing them to better track their progress. 
\section {Method}
\label{sec:method}
\subsection{Metrics}
The evaluation of unsupervised learning methods is a complicated topic, and many different metrics were developed. In this section, we briefly review the metrics we utilized for clustering evaluation.
\paragraph{Accuracy}
In the presence of ground truth labels, it is possible to evaluate the prediction accuracy by assigning classes to predicted clusters. 
Similarly to previous works \cite{xie2015dec,jiang2016variational,gansbeke2020scan} we use linear assignment \cite{kuhn1955hungarian,crouse2016implementing} for assignment of clusters to the classes. In cases when the number of clusters is larger than the number of classes (overclustering), we assign one cluster to each class, while the rest is assigned greedily to maximize accuracy.
\paragraph{Normalized Mutual Information (V-measure)}
For a partition of the instances, $U$, we define entropy as
\begin{align}
    H(U)=-\sum_{i=1}^R P_U(i)\log(P_U(i)),
\end{align}
and mutual information between two partitions as
\begin{align}
    \mathrm{MI}(U,V)=\sum_{i=1}^R \sum_{j=1}^C P_{UV}(i,j)\log (\frac{P_{UV}(i,j)}{P_U(i)P_V(j)}),
\end{align}
where
\begin{align}
    P_{UV}(i,j) &= \frac{\abs{U_i}\abs{V_j}}{N}\\
    P_U(i) &= \frac{\abs{U_i}}{N}.
\end{align}
To be able to compare mutual information in different cases, it is usually normalized \cite{kvalseth1987entropy}:
\begin{align}
    \mathrm{NMI}(U,V) = \frac{\mathrm{MI}(U,V)}{\mathrm{avg}(H(U), H(V))},
\end{align}
where $\mathrm{avg}$ is some function, in our case the arithmetic mean.
\paragraph{Adjusted Mutual Information}
Since mutual information tends to have larger values when the number of clusters is large,  mutual information should be adjusted for random chance \cite{vinh2010information}
\begin{align}
    \mathrm{AMI}(U, V) = \frac{\mathrm{MI}(U, V) - \mathbb{E}\qty[\mathrm{MI}(U, V)]}{[\mathrm{avg}(H(U), H(V)) - \mathbb{E}\qty[MI(U, V)]}.
\end{align}
\paragraph{Adjusted Rand Index }
Rand index \cite{rand1971objective} is another measure of clustering quality. It can be viewed as an accuracy measure over pairs of instances: denoting the number of pairs as $N_p = \binom{N}{2}$, the number of pairs of instances that belong to the same set in both partitions as TP, and the number of pairs of instances that belong to the different sets in both partitions as TN, we define Rand index as
\begin{align}
    \mathrm{RI} = \frac{\mathrm{TP} + \mathrm{TN}}{N_p}.
\end{align}
We also adjust the index for chance in the usual manner \cite{hubert1985comparing}:
\begin{align}
    \mathrm{ARI} = \frac{\mathrm{RI}(U, V) - \mathbb{E}\qty[\mathrm{RI}(U, V)]}{[1 - \mathbb{E}\qty[\mathrm{RI}(U, V)]},
\end{align}
where $1$ is the maximal value of Rand index. 

\subsection{Evaluation}
To train a clustering model, we extract features of both the training and the validation set with a pre-trained model. We do not apply any augmentations during feature extraction. As opposed to the existing clustering approaches, e.g., DeepCluster \cite{caron2018deepclustering}, our method does not utilize a clustering objective as a part of feature extractor training, but merely uses a feature extractor pre-trained in a self-supervised manner.

Modern clustering approaches are usually based on some distance between different samples. Unfortunately, if the dimension of space is high, the distance between samples provides a little information. In our case, since most of the methods provide at least $1000$-dimensional embeddings, we apply dimensional reduction. In particular, we train incremental PCA model with batch size $\max\qty(4096, 2\cdot n_f)$, where $n_f$ is the dimension of extracted features. 

After applying dimensional reduction, we train mini-batch variation of k-means with the transformed features. Since the features are extracted only once, the training clustering model is relatively cheap. Depending on the model, it takes a couple of hours on CPU.  Using augmentation and training first PCA and then clustering models can boost performance but is much more resource-demanding.

While by default, we set the number of clusters to be $1000$ (number of ImageNet classes), we also experiment with overclustering, following \citet{gansbeke2020scan}.
\section{Experimental Results}
\label{sec:exp}

We evaluate recent state-of-the-art approaches with the proposed protocol: MoCo v2 \cite{chen2020moco2}, InfoMin \cite{tian2020infomin}, SwAV \cite{caron2020swav}, SimCLRv2 \cite{chen2020simclr2}, and BigBiGAN \cite{donahue2019bigbigan}. 
For every paper, we evaluate ResNet-50 and best performing network. 
We also add results for three models trained in supervised manner\footnote{We employ evaluation code by \citet{wightman2020models}.}: ResNet-152,  EfficientNet-L2 \cite{xie2019noisystudent}, and IG-ResNeXt-101 32$\times$48d 
 \cite{mahajan2018weaklyig}.

For experiments, we utilize feature extracted from two different datasets: ImageNet and ObjectNet \cite{barbu2019objectnet}.

We trained k-means for 60 epochs, but even 1 epoch often gets decent results.

\begin{figure*}
	\centering
	\begin{subfigure}[b]{0.48\linewidth}
		\includegraphics[width=\linewidth]{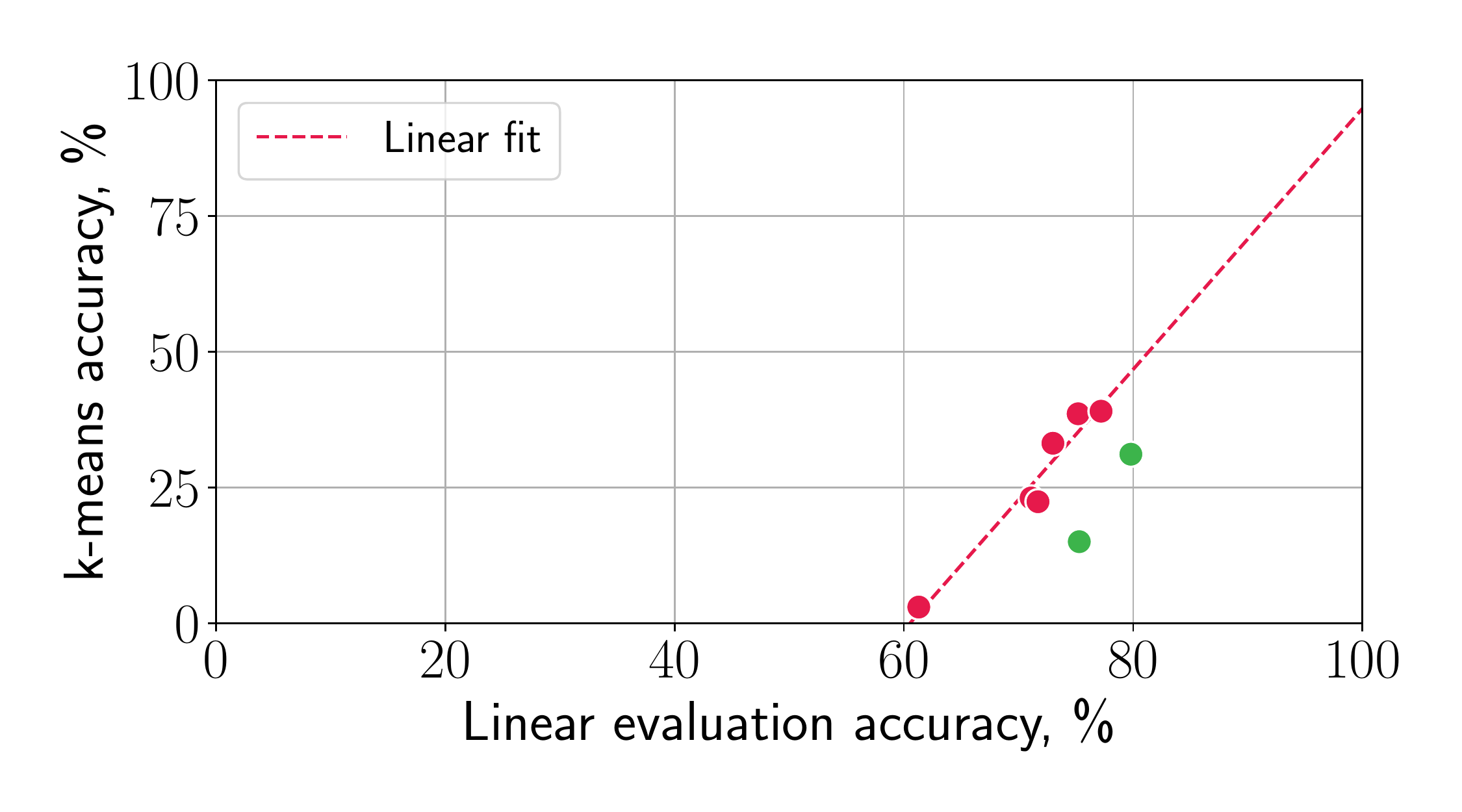}
		\subcaption{ }
		\label{fig:lin_vs_uns}
	\end{subfigure}
	\begin{subfigure}[b]{0.48\linewidth}
		\includegraphics[width=\linewidth]{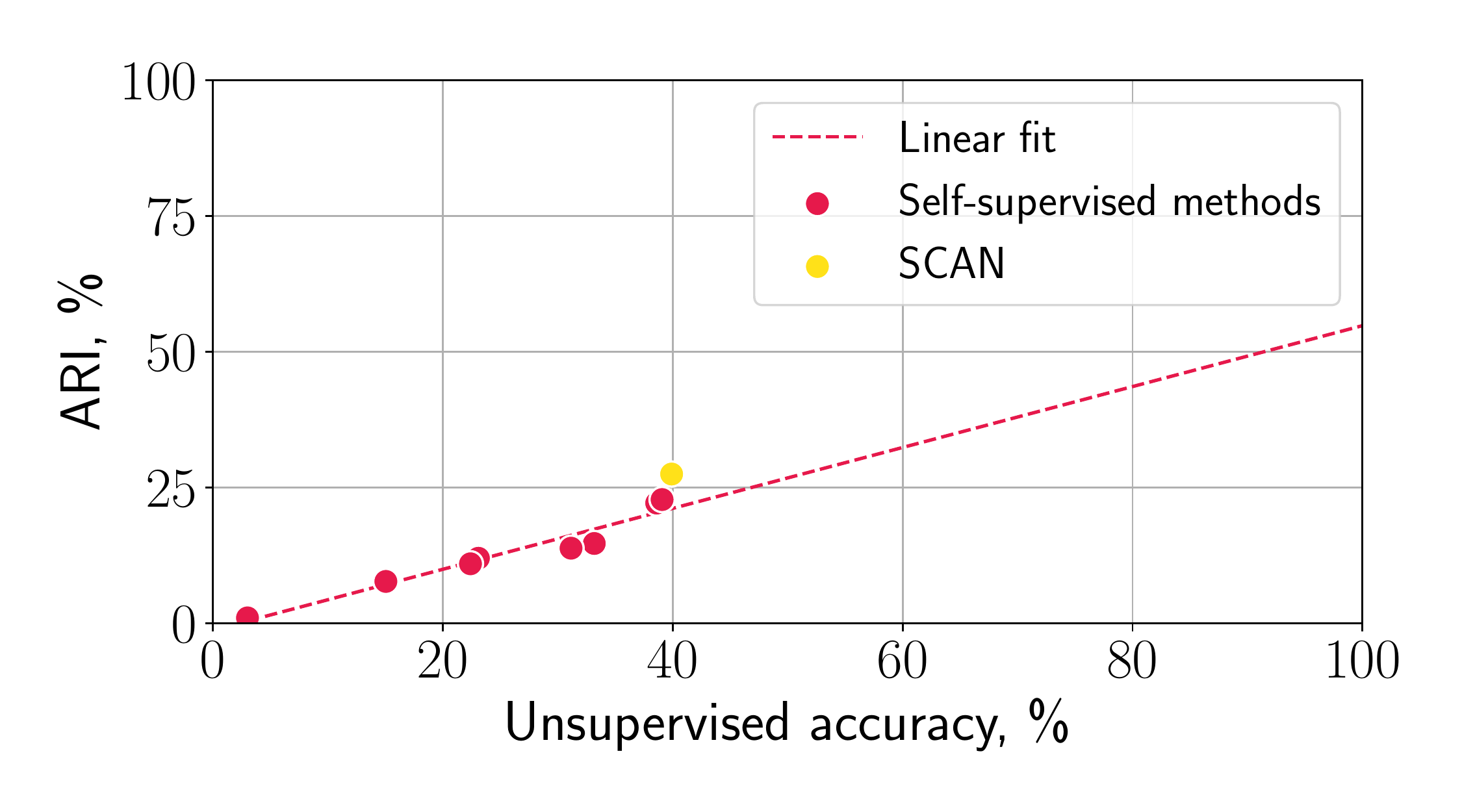}
		\subcaption{ }
		\label{fig:uns_vs_ari}
	\end{subfigure}
	\caption{ Visualization of different metrics: (a) unsupervised accuracy and  linear evaluation accuracy; (b) ARI and unsupervised accuracy. Green points are outliers (SimCLRv2 (ResNet-152 3$\times$, SK) and SwAV).
	}
	\label{fig:metrics}
\end{figure*}

\paragraph{ImageNet}
Experimental results for ImageNet are shown in \cref{tab:res_inet}. During accuracy calculation, we used training labels for cluster assignment. In addition, we visualize different metrics in \cref{fig:metrics}. We note a strong correlation between linear evaluation accuracy and k-means accuracy, except for SimCLRv2 (ResNet-152 3$\times$, SK) and SwAV. We note that SCAN \cite{gansbeke2020scan} gives significantly larger ARI for similar accuracy values. We also note that both supervised and self-supervised methods with high-dimensional embeddings ( ResNet-152 3$\times$ and EfficientNet-L2) show weaker results than their counterparts.

\begin{table}
\centering
\caption{Experimental results on ImageNet in form mean$\pm$std of 5 runs. Overclustering denoted as ``over.'', supervised models denoted as ``super.'' Bold denotes highest results among our experiments, and red denotes results within one standard deviation of best results. Results for self-label are taken from the paper's official repository.} 
\label{tab:res_inet}
\resizebox{\linewidth}{!}{
\begin{tabular}{lccccc}
\toprule
\textbf{Method}   & \textbf{Linear (super.)}& \textbf{ACC}& \textbf{ARI}& \textbf{AMI}& \textbf{NMI}\\ \midrule
MoCo v2 (ResNet-50)&$71.1$&$23.09\pm0.16$&$11.99\pm0.13$&$37.04\pm0.10$&$63.22\pm0.05$\\ 
InfoMin (ResNet-50)&$73.0$&$33.17\pm0.32$&$14.71\pm0.38$&$48.25\pm0.27$&$68.80\pm0.17$\\ 
SwAV (ResNet-50)& $75.3$ & $15.04 \pm 0.77$ & $7.72 \pm 0.33$ & $ 32.33 \pm 0.16$&  $55.34 \pm 0.62 $\\ 
SimCLRv2 (ResNet-50)&$71.7$&$22.40\pm0.19$&$10.97\pm0.20$&$34.85\pm0.29$&$61.52\pm0.18$\\ \midrule
BigBiGAN (RevNet-50 4$\times$)&$61.3$&$3.00\pm0.09$&$1.01\pm0.04$&$8.81\pm0.27$&$35.99\pm0.69$\\ 
InfoMin (ResNeXt-152)&$75.2$&\textcolor{red}{$38.60\pm0.67$}&$22.15\pm 0.52$&$\bm{52.56\pm0.11}$&$\bm{72.17\pm 0.13 }$\\ 
SimCLRv2 (ResNet-152, SK)&$77.2$&$\bm{39.07 \pm 0.61}$&$\bm{22.80\pm 0.60}$&$52.03\pm 0.19$&$71.83\pm 0.13$\\ 
SimCLRv2 (ResNet-152 3$\times$, SK)&$\bm{79.8}$&$31.15\pm0.74$&$13.84\pm0.84$&$46.64\pm0.25$&$65.79\pm0.58$\\\midrule
SimCLRv2 (ResNet-152, SK, 1.5$\times$ over.)&$77.2$&$46.03 \pm 
0.21$&$23.94 \pm  0.16$&$50.77 \pm 0.25$&$73.14 \pm 0.06$\\ \midrule
SCAN \cite{gansbeke2020scan}&$-$&$39.9$&$27.5$&$51.2$&$72.0$\\   
Self-label \cite{asano2019selflabelling}&$63.5$&$30.5$&$16.2$&$42.0$&$75.4$\\ 
Self-label $3\times$ over. \cite{asano2019selflabelling}&$68.8$&$38.1$&$27.6$&$52.8$&$75.7$\\ \midrule
ResNet-152 (super.) &$81.0$&$65.60\pm 0.93$&$53.02\pm0.76$&$74.02\pm0.22$&$84.97\pm0.17$\\  
IG-ResNeXt-101 32$\times$48d  (super.) &$85.4$&$72.39\pm0.52$&$63.31 \pm0.40$& $81.17 \pm 0.08$ &$89.23\pm 0.05$\\  
EfficientNet-L2 (super.) &$88.2$&$59.08 \pm 0.67$& $46.32\pm 0.60$ &$69.35 \pm
0.26$ &$82.33 \pm 0.18$\\  
 \bottomrule
\end{tabular}}
\end{table}

\paragraph{ObjectNet}
To access the generalization of acquired clustering, in addition to ImageNet, we evaluate the proposed method on ObjectNet. ObjectNet is a test set for vision tasks, created to control the performance of vision algorithms in settings close to real life.

\cref{tab:res_onet} shows results for k-means trained on ImageNet training set. In this case, we evaluated only on intersecting classes between ImageNet and ObjectNet. We show accuracy both for cluster assignment based on ImageNet training set (ACC-tr) and ObjectNet itself (ACC-val). Note that since some ObjectNet classes are mapped to two different ImageNet classes, assigning all images to a single class will result in  ${\sim}1.77\%$ accuracy. By manually inspecting the predictions of the k-means, we conclude that in many cases, assignment of a large part of instances to a single class indeed happens.

When the ImageNet cluster assignment is used, no network, including supervised ones, show better-than-random performance. 
For ObjectNet assignment, only BigBiGAN (as well as supervised networks) is significantly better than assigning all the instances to a single class. Moreover, ResNet-50 shows better results than larger networks among different self-supervised networks. We advise that, as for now, AMI should be used as a metric for tracking progress on this task.

\cref{tab:res_onet_tr} shows results for clustering trained directly on ObjectNet.  For pre-trained models, the performance on classes that are part of ImageNet is much better: for example, IG-ResNeXt-101 32$\times$48d, has $41.36\%$ accuracy as compared to $15.81\%$ for classes not in ImageNet. For self-supervised, the difference is much smaller: for InfoMin, performance on classes not included in ImageNet is the same ($6.53\%$).

\begin{table}
\centering
\caption{
Experimental results on ObjectNet in form mean$\pm$std of 5 runs, using clusters acquired from ImageNet training. }
\label{tab:res_onet}
\resizebox{\linewidth}{!}{
\begin{tabular}{lccccc}
\toprule
\textbf{Method}   &  \textbf{ACC-tr}& \textbf{ACC-val}& \textbf{ARI}& \textbf{AMI}& \textbf{NMI}\\ \midrule
MoCo v2 (ResNet-50) &$0.11\pm0.19$&$1.76\pm0.20$&$0.02\pm0.04$&$0.42\pm0.49$&$0.82\pm0.66$\\ 
InfoMin (ResNet-50) &$0.12\pm0.26$&$2.18\pm0.25$&$0.08\pm0.07$&$\bm{1.81\pm0.57}$&$2.34\pm0.56$\\ 
SwAV (ResNet-50)&$0.21 \pm 0.33$ &$1.85 \pm  0.12$ &$0.06\pm0.08$&$0.65\pm0.34$&$1.30\pm0.48$ \\
SimCLRv2 (ResNet-50)&$0.00\pm0.00$&$2.14\pm0.30$&$0.06\pm0.06$&\textcolor{red}{$1.47\pm0.76$}&$2.43\pm0.75$\\ \midrule
BigBiGAN (RevNet-50 4$\times$) &$0.10\pm0.01$&$\bm{4.92\pm0.20}$&$ \bm{0.10\pm0.01}$&$1.00\pm0.06 $&$\bm{15.98\pm0.69 }$\\ 
InfoMin (ResNeXt-152) &$\bm{0.67\pm0.39}$&$1.96\pm0.39$&$0.01\pm0.01$&$0.70\pm0.37$&$1.26\pm0.58$\\ 
SimCLRv2 (ResNet-152, SK)&$0.00\pm 0.00$&$1.69\pm0.28$&$0.03\pm0.07$&$0.55\pm 0.82$&$0.86\pm 0.94 $\\ 
SimCLRv2 (ResNet-152 3$\times$, SK)&$0.00\pm 0.00$&$1.72\pm0.19$&$0.01\pm0.01$&$0.44\pm0.42$&$0.94\pm0.61$\\   \midrule
SimCLRv2 (ResNet-152, SK, 1.5$\times$ over.)&$0.04 \pm
0.10$&$1.75 \pm 0.19$&$0.01\pm0.01$&$0.45\pm0.38$&$0.99\pm0.50$\\ \midrule
ResNet-152 (super.) &$0.36\pm0.48$& $1.75\pm
0.35$&$0.03\pm0.07$&$0.53 \pm0.97$&$0.76\pm 1.19$\\ 
IG-ResNeXt-101 32$\times$48d  (super.) &$0.04 \pm 0.08$& $2.15 \pm 0.84$ &$0.14\pm 0.21$&$2.12 \pm 2.91$&$2.51 \pm 3.31$\\ 
EfficientNet-L2 (super.) &$0.36 \pm 0.44$& $2.10 \pm0,41$ &$0.13\pm0.14$&$1.95 \pm 1.37$&$2.34\pm 1.51$\\  
 \bottomrule
\end{tabular}}
\end{table}

\begin{table}
\centering
\caption{Experimental results  on ObjectNet using clusters acquired by training k-means on ObjectNet itself.}
\label{tab:res_onet_tr}
\resizebox{\linewidth}{!}{
\begin{tabular}{lcccc}
\toprule
\textbf{Method}   &  \textbf{ACC}& \textbf{ARI}& \textbf{AMI}& \textbf{NMI}\\ \midrule
MoCo v2 (ResNet-50)&$4.30\pm0.05$&$0.77\pm0.02$&$8.08\pm0.10$&$20.57\pm0.39$\\ 
InfoMin (ResNet-50) &$4.96\pm 0.08$&$0.92\pm0.22$&$8.85\pm0.08$&$21.49\pm0.17$\\ 
SwAV (ResNet-50)&$3.44\pm0.11$ &$0.60\pm0.04$ &$6.77\pm0.11$ &$16.20\pm 0.54 $\\ 
SimCLRv2 (ResNet-50)&$3.67\pm0.22$&$0.63\pm0.02$&$6.72\pm0.09$&$18.75\pm 0.24 $\\  \midrule
BigBiGAN (RevNet-50 4$\times$) &$2.30\pm0.03$&$0.116\pm0.001$&$1.75\pm0.03$&$14.93\pm0.22$\\ 
InfoMin (ResNeXt-152) &$\bm{6.53\pm 0.19}$&$\bm{1.59\pm0.04}$&$\bm{12.49\pm 0.13}$&$\bm{24.97\pm0.24}$\\ 
SimCLRv2 (ResNet-152, SK) &$5.34\pm 0.20$&$1.15\pm 0.07$&$9.24\pm0.23$&$22.08\pm0.17$\\ 
SimCLRv2 (ResNet-152 3$\times$, SK) &$4.20\pm0.19$&$1.00\pm0.08$&$8.62\pm0.16$&$17.53\pm0.27$\\  \midrule
SimCLRv2 (ResNet-152, SK, 1.5$\times$ over.) &$6.47 \pm 0.07$ &$1.32 \pm
0.05$ &$9.46 \pm 0.08$ &$23.62 \pm 0.28$\\ \midrule
ResNet-152 (super.) &$14.36\pm
1.80$&$6.09\pm 1.20$&$23.93\pm 0.26$&$32.60\pm 2.55$ \\ 
IG-ResNeXt-101 32$\times$48d (super.) &$25.25 \pm
0.46$&$14.03 \pm 0.18$&$36.30 \pm 0.11$ &$44.72 \pm 0.24$ \\ 
EfficientNet-L2 (super.) &$7.70 \pm 0.57$&$2.07 \pm 0.30$&$17.78 \pm 0.44$&$22.60 \pm 0.61$ \\ 
 \bottomrule
\end{tabular}}
\end{table}
\subsection{Ablation study}

\begin{figure*}
	\centering
	\begin{subfigure}[b]{0.48\linewidth}
		\includegraphics[width=\linewidth]{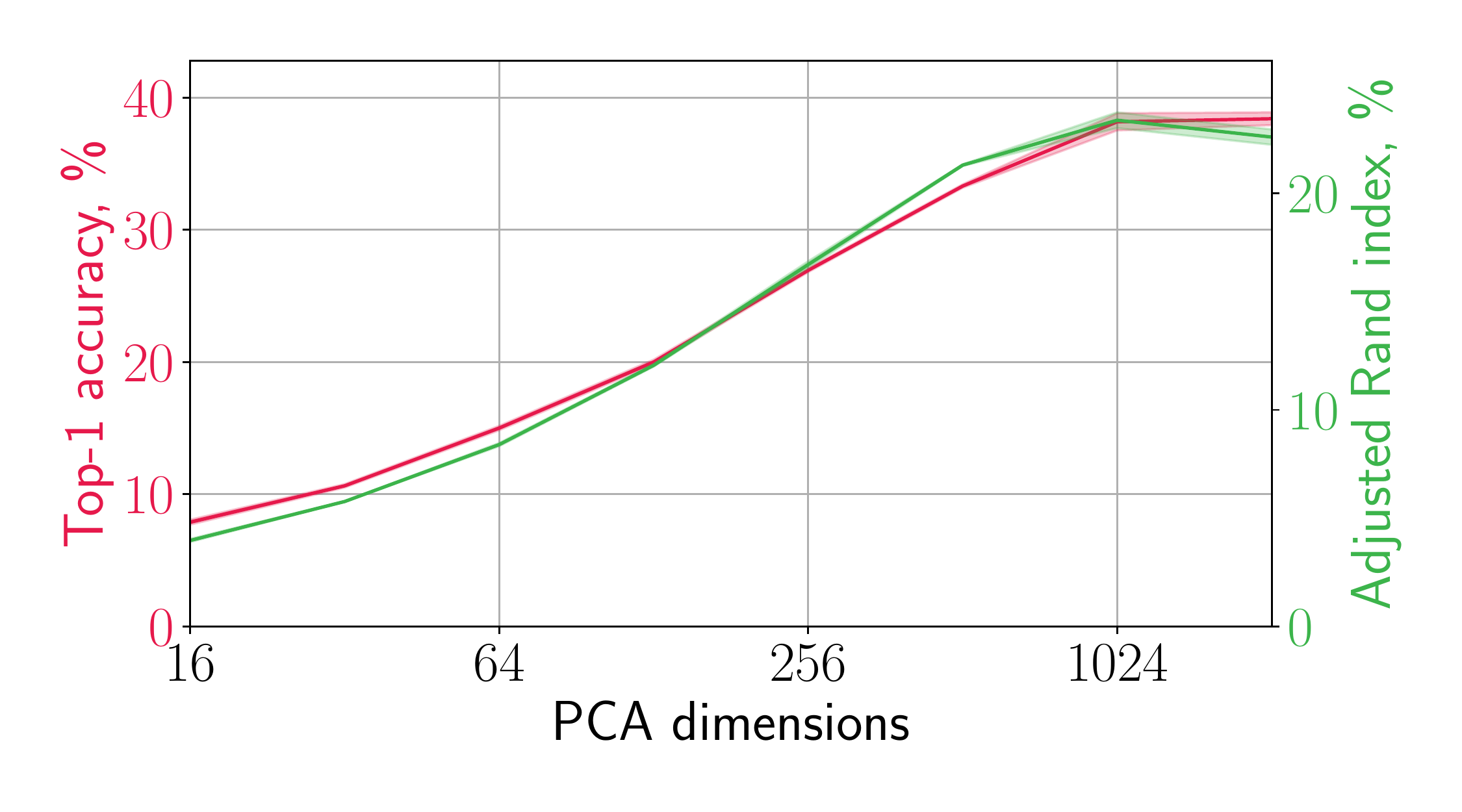}
		\subcaption{Accuracy and ARI as a function of dimensions after dimensional reduction.}
		\label{fig:pca}
	\end{subfigure}
	\begin{subfigure}[b]{0.48\linewidth}
		\includegraphics[width=\linewidth]{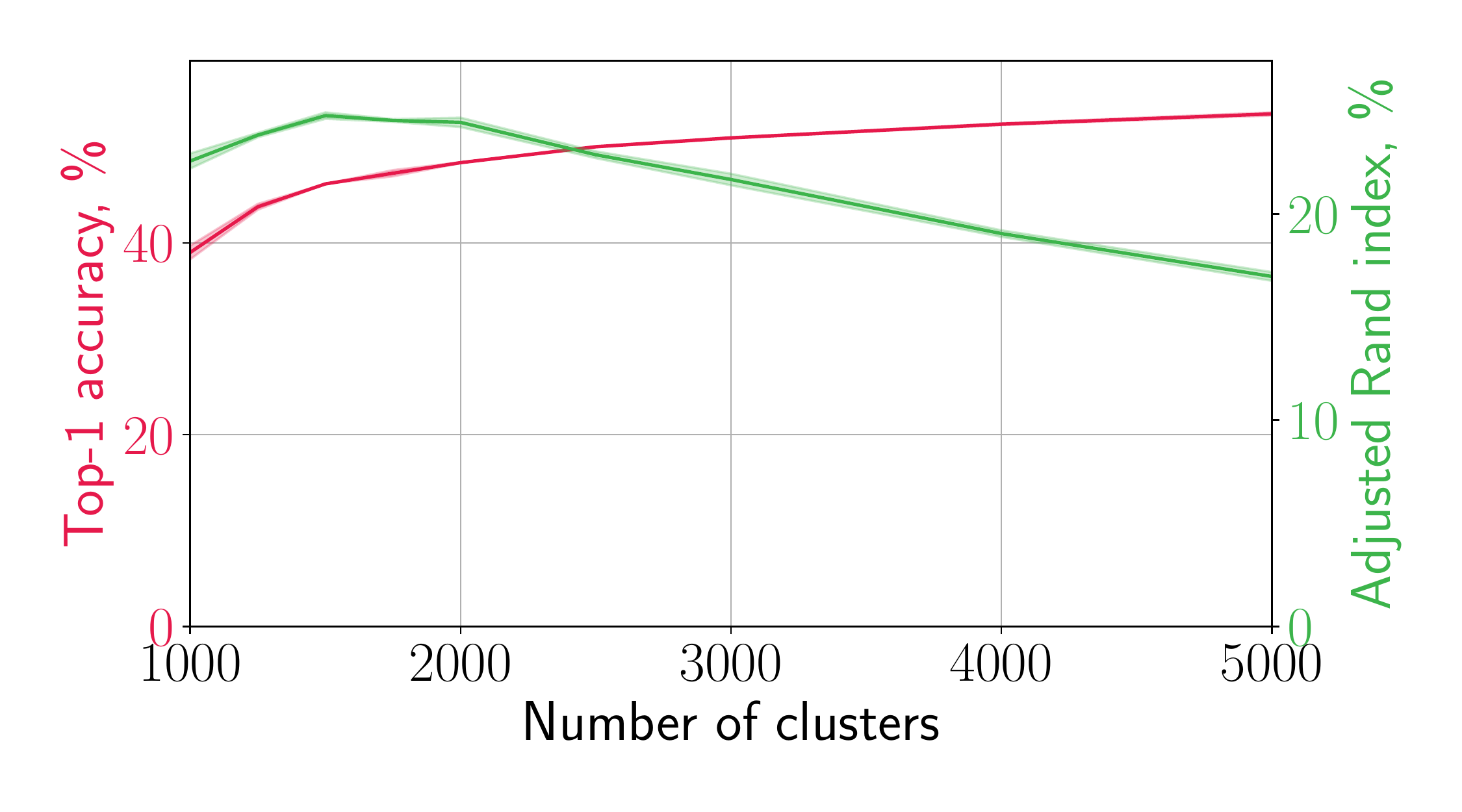}
		\subcaption{Accuracy and ARI as a function of number of clusters. }
		\label{fig:over}
	\end{subfigure}
	\caption{Ablation study for the best-performing model, SimCLRv2 (ResNet-152, SK).
	}
	\label{fig:ablation}
\end{figure*}

\paragraph{Dimensionality reduction}
\cref{fig:pca} shows the effect of the number of dimensions used for clustering. As expected, an increasing number of dimensions provide diminishing returns and might harm the results for more than 1024 dimensions.
\paragraph{Overclustering}
Since some classes in ImageNet may contain fairly different images, increasing the number of clusters beyond 1000 improves not only accuracy (since this metric uses real labels, its calculation inevitably involves passing information), but also ARI, as shown in \cref{fig:over}. For that reason, we add $1.5 \times$ overclustering version of the best-performing model to comparison.
\section{Conclusion}
\label{sec:conclusion}
In this paper, we study the applications of self-supervised learning for unsupervised classification. We establish competitive baselines by just applying PCA dimensional reduction and k-means clustering to features extracted by existing self-supervised methods. Thanks to a practice of publishing both code and pre-trained models, we were able to evaluate multiple state-of-the-art approaches and achieve as high as $39\%$ accuracy on ImageNet in an unsupervised manner and $46\%$ with overclustering. 

Also, we propose an unsupervised clustering of extracted features as an additional way to evaluate self-supervised training approaches, along with linear evaluation and transfer learning.

Finally, we raise several issues and possible directions for future work. First, the question of whether severe underperformance of models with higher-dimensional feature space, such as ResNet $3\times$, remains open. Is it the weakness of the proposed clustering method or rather a property of the model? Can the reduction of the dimension of the embeddings improve performance on other tasks?  

Second, the models' poor performance transferred to ObjectNet, even among supervised models, with a prominent exception of BigBiGAN, is of great interest. Is it possible to achieve high performance on ObjectNet and ImageNet simultaneously, at least at the linear classification level? What is the reason BigBiGAN is the only model showing better-than-random results on ObjectNet? What is performance on other ImageNet-related datasets such as ReaL labels \cite{beyer2020arewedone}, ImageNetV2 \cite{recht19imagenetv2}, ImageNet-R \cite{hendrycks2020imagenetr}, etc.?

Last, can the approach itself be improved? Can we take into account the method during the self-supervised training without significant performance degradation in other tasks? What is the better approach for dimensional reduction and clustering itself? What is the effect of augmentation in the clustering training phase?

We hope this paper will raise an interest in self-supervised approach to large-scale image clustering.  



\clearpage
%
%
\bibliographystyle{plainnat}
\bibliography{uns_eccv}

\newpage

\appendix{}

\renewcommand\thefigure{\thesection.\arabic{figure}} 
\renewcommand\thetable{\thesection.\arabic{table}} 
\renewcommand\theequation{\thesection.\arabic{equation}}  
\setcounter{figure}{0}  
\setcounter{table}{0}

\crefalias{section}{appsec}
\crefalias{subsection}{appsec}
\crefalias{subsubsection}{appsec}

\end{document}